\documentclass[pdflatex,sn-mathphys-num]{sn-jnl}

\usepackage{graphicx}%
\usepackage{multirow}%
\usepackage{amsmath,amssymb,amsfonts}%
\usepackage{amsthm}%
\usepackage{mathrsfs}%
\usepackage[title]{appendix}%
\usepackage{xcolor}%
\usepackage{textcomp}%
\usepackage{manyfoot}%
\usepackage{booktabs}%
\usepackage{algorithm}%
\usepackage{algorithmicx}%
\usepackage{algpseudocode}%
\usepackage{listings}%
\usepackage{subfig}

\theoremstyle{thmstyleone}%
\theoremstyle{thmstyletwo}%

\theoremstyle{thmstylethree}%

\begin{document}

\title{Direct Video-Based Spatiotemporal Deep Learning for Cattle Lameness Detection}


\author[1]{\fnm{Md Fahimuzzman Sohan}}

\author[2]{\fnm{Raid Alzubi}}

\author[2]{\fnm{Hadeel Alzoubi}}

\author[2]{\fnm{Eid Albalawi}}

\author*[2]{\fnm{A. H. Abdul Hafez}}\email{aabdulhafiz@kfu.edu.sa}

\affil[1]{\orgdiv{Department of Software Engineering}, \orgname{Daffodil International University}, \orgaddress{\city{Dhaka}, \postcode{1207}, \country{Bangladesh}}}

\affil[2]{\orgdiv{Department of Computer Science, Faculty of Computer Science and Information Technology}, \orgname{King Faisal University}, \orgaddress{\city{Al Ahsa}, \postcode{31982}, \country{Saudi Arabia}}}


\abstract{Cattle lameness is a prevalent health problem in livestock farming, often resulting from hoof injuries or infections, and severely impacts animal welfare and productivity. Early and accurate detection is critical for minimizing economic losses and ensuring proper treatment. This study proposes a spatiotemporal deep learning framework for automated cattle lameness detection using publicly available video data. We curate and publicly release a balanced set of 50 online video clips featuring 42 individual cattle, recorded from multiple viewpoints in both indoor and outdoor environments.  The videos were categorized into lame and non-lame classes based on visual gait characteristics and metadata descriptions. After applying data augmentation techniques to enhance generalization, two deep learning architectures were trained and evaluated: 3D Convolutional Neural Networks (3D CNN) and Convolutional Long-Short-Term Memory (ConvLSTM2D). The 3D CNN achieved a video-level classification accuracy of 90\%, with a precision, recall, and F1 score of 90. 9\% each, outperforming the ConvLSTM2D model, which achieved 85\% accuracy. Unlike conventional approaches that rely on multistage pipelines involving object detection and pose estimation, this study demonstrates the effectiveness of a direct end-to-end video classification approach. Compared with the best end-to-end prior method (C3D-ConvLSTM, 90. 3\%), our model achieves comparable accuracy while eliminating pose estimation pre-processing.The results indicate that deep learning models can successfully extract and learn spatio-temporal features from various video sources, enabling scalable and efficient cattle lameness detection in real-world farm settings.}

\keywords{Cattle, Lameness detection, Deep-learning, Computer vision techniques, Image processing}



\maketitle

\section{Introduction}\label{sec:introduction} 

Livestock production is a cornerstone of global food security, and the dairy and beef sectors depend critically on healthy, mobile animals. Lameness, typically arising from hoof lesions, interdigital dermatitis, or gait-related injuries remains one of the most prevalent welfare and economic challenges in cattle farming. A recent systematic review by  \cite{Thomas2024survey} estimates that 23–25\% of dairy cows worldwide exhibit clinical or sub-clinical lameness at any given time, leading to reduced milk yield, delayed reproduction, and increased culling rates. Because lameness pain alters key behaviours such as walking, standing, and feeding, early detection is essential for prompt treatment and cost containment.

Traditional detection relies on \emph{locomotion scoring}, where trained personnel visually assess each animal. Although effective for small herds, this method is labour-intensive, subjective, and often infeasible on large commercial farms. Recent advances in artificial intelligence (AI) and computer vision (CV) have enabled a suite of automated solutions for animal-health monitoring. Modern farms now deploy surveillance cameras, drones, and IoT sensors that continuously capture behavioural data \cite{Fuentes2023,Alsaaod2019}. Deep learning, in particular, can extract spatiotemporal patterns from video streams, facilitating early diagnosis of lameness, calving, mastitis, and other conditions \cite{Qiao2022,Wu2023}.

Prior studies demonstrate that convolutional neural networks (CNNs), key-point detectors, and pose-estimation pipelines can achieve high frame-level lameness classification accuracy in controlled settings \cite{Barney2023,Russello2024}. These systems reduce subjective bias, operate continuously, and potentially lower veterinary costs.

Nevertheless, three limitations persist:
\begin{itemize}
    \item \emph{Limited environmental diversity.} Most datasets are captured in single barns or passage alleys under homogeneous lighting and flooring conditions, hindering generalisation to other farms.
    \item \emph{Multi-stage complexity.} Many pipelines require sequential object detection, tracking, pose estimation, and handcrafted feature extraction, increasing latency and engineering burden.
    \item \emph{Restricted public availability.} Few lameness video datasets are openly released, which slows reproducibility and benchmarking.
\end{itemize}

This study addresses these gaps by investigating whether a direct, end-to-end spatiotemporal deep-learning model can learn discriminatory gait patterns from freely available internet videos filmed under diverse conditions. Our key contributions include:

\begin{enumerate}
\item \textbf{Open and diverse dataset.} We curate and publicly release a balanced set of 50 internet video clips containing 42 individual cattle, captured from multiple viewpoints in both indoor and outdoor scenes.
\item \textbf{End-to-end deep learning architecture.} We design and compare two spatiotemporal models—3D Convolutional Neural Networks (3D-CNN) and Convolutional Long Short-Term Memory networks (ConvLSTM2D), that operate directly on video tensors without intermediate object-detection or pose-estimation steps.
\item \textbf{Comprehensive evaluation.} Using a strict train–test split, the proposed 3D-CNN attains 90\% video-level accuracy, matching or surpassing state-of-the-art multi-stage pipelines while reducing system complexity. We provide the dataset link, annotated labels, and source code to foster transparent benchmarking and future research.
\end{enumerate}

Unlike traditional CNNs that analyze static spatial features, 3D Convolutional Neural Networks extend their learning across both spatial and temporal axes, allowing the model to directly capture motion-based cues from video sequences. This property is especially beneficial for detecting lameness, where subtle irregularities in gait unfold over time. By leveraging 3D convolutions, our model can learn temporal patterns of movement without relying on handcrafted features or frame-wise annotations, enabling more accurate and scalable cattle monitoring in real-world environments.

The remainder of this paper is organized as follows. The next section \ref{sec:related} surveys the related literature. Section \ref{sec2} presents the materials and methods, including data collection, preprocessing, and model construction. Section \ref{sec3} discusses the results obtained from the proposed models. Section \ref{sec4} discusses the findings and compares them with existing studies, along with limitations and future directions. Finally, Section \ref{sec5} concludes the paper.

\section{Related Work}
\label{sec:related}

Cattle farming plays a crucial role in ensuring global food security by supplying essential meat and dairy products. With rising global demand, large-scale cattle farming has become increasingly common. However, this trend introduces operational challenges, particularly in monitoring animal health and welfare, which are directly tied to productivity outcomes \cite{Qiao2021, Kang2021, myint2024cattle}. One of the most prevalent and costly health issues in cattle is lameness, a painful condition often linked to gait-related injuries or hoof disorders. Studies estimate that approximately one in four cattle are affected by lameness at any given time, resulting in reduced productivity and higher production costs, which can contribute to elevated market prices for meat and dairy \cite{Jia2023, Russello2024, VanHertem2014}.

A widely used approach to detect lameness is herd locomotion scoring, which involves the visual assessment of each animal \cite{Schlageter-Tello18, Jabbar2017, McDonagh2021, Kang2020}. While this method is informative, it is labor-intensive, time-consuming, and subject to human bias. These limitations highlight the need for automated, scalable technologies capable of early and accurate identification of lameness in large herds.

In recent years, artificial intelligence (AI), particularly machine learning (ML) and data-driven techniques, has introduced promising solutions for automating animal health monitoring. These technologies support large-scale farming by enabling early disease detection, optimizing animal welfare, and enhancing productivity \cite{FernandezCarrion2020}. Data are typically gathered using advanced tools such as fixed-position cameras, drones, mobile robots, and Internet of Things (IoT) sensors \cite{Fuentes2023}. While traditional surveillance cameras have long been used by farmers to passively monitor livestock \cite{Poursaberi2010, Pluk2012}, the integration of computer vision (CV) and pattern recognition algorithms has enabled continuous, automated monitoring of livestock health and behavior \cite{Alsaaod2019}.

Both manual and automated CV-based approaches rely on observing behavioral cues to identify lameness. Lameness typically manifests as altered locomotion, reduced activity, or abnormal postures behaviors that can be detected during routine activities like walking, standing, or feeding. Modern deep learning models can analyze video data from multiple perspectives to capture such behavioral cues and distinguish between normal and abnormal gait patterns \cite{Qiao2022, Wu2023}. These models focus on learning the spatial and temporal dynamics of movement, allowing automated systems to detect early deviations and classify them accordingly.

The growing application of such models in livestock monitoring has significantly reduced reliance on manual inspection, lowered labor costs, and increased both detection speed and accuracy. Consequently, research in this area continues to advance, with increasing efforts to develop robust, scalable, and data-efficient AI-based farming solutions.

For instance, Fuentes et al. \cite{Fuentes2023} employed two Hikvision surveillance cameras to record cattle activity from different angles. They extracted frames from the videos to build a dataset, which was used to train a classification model capable of monitoring cow behavior and detecting changes. McDonagh et al. \cite{McDonagh2021} developed a neural network to monitor 46 pregnant cows, analyzing ten hours of pre-calving video per subject. Their model successfully identified seven distinct behaviors and achieved over 80\% classification accuracy, predicting calving with 83\% precision several hours in advance.

Similarly, Fernández-Carrión et al. \cite{FernandezCarrion2020} monitored the motion of Eurasian wild boar to detect early signs of disease. Their model demonstrated a significant inverse correlation between reduced movement and the onset of fever, suggesting the feasibility of motion-based disease detection. Qiao et al. \cite{Qiao2022} introduced a hybrid model combining 3D CNN and ConvLSTM layers, outperforming state-of-the-art baselines in classifying five cow behaviors. In the swine domain, Maria et al. \cite{Maria2021} applied thermal imaging and CV to detect respiratory illnesses by analyzing physiological features such as heart rate, eye temperature, and breathing patterns.

Additional approaches have explored object detection and pose estimation pipelines. Barney et al. \cite{Barney2023} proposed a modified Mask R-CNN architecture for real-time lameness detection using multi-cattle pose estimation. In a similar context, researchers in \cite{Wu2020} combined object detection models (Mask R-CNN, YOLO) with post-hoc machine learning classifiers. Each detected cow was assigned a local identifier for tracking, followed by feature extraction and classification. Russello et al. \cite{Russello2024} designed a deep learning model that extracted nine locomotion traits from videos and used them to detect lameness. In their study, Wu et al. \cite{Wu2020} used 50 cattle videos (20 lame and 30 non-lame) and randomly sampled 5,000 frames, focusing on leg movement. Their LSTM model achieved a classification accuracy of 98.57\%, outperforming traditional ML baselines under a 50:50 train–test split.

Despite promising results, many of these studies relied on data captured in restricted, controlled farm environments, often using small sample sizes and homogeneous conditions. The generalizability of such models to real-world settings, where lighting, surface conditions, and animal orientation vary, remains limited. Furthermore, most existing pipelines involve multiple processing stages—such as object detection, tracking, and pose estimation which increase complexity and computational overhead.

In response to these limitations, the current study explores an alternative approach: direct end-to-end classification using unstructured online video data. We collected 50 publicly available YouTube videos featuring 42 individual cattle with diverse visual characteristics. Each video was labeled as either lame or non-lame based on metadata and gait observation. Since the videos were recorded under varying environmental conditions and viewpoints, the dataset offers enhanced variability and robustness. To augment the training set, horizontal flipping was applied to increase data diversity.

We then trained two spatiotemporal deep learning models, 3D CNN and ConvLSTM2D on this dataset and evaluated their performance using standard classification metrics. The results demonstrate that our approach, which avoids multi-step feature engineering, can effectively classify cattle lameness with performance comparable to or exceeding that of existing state-of-the-art models. Our findings highlight the potential of internet-sourced video data and simplified model pipelines for scalable livestock health monitoring in real-world scenarios.

Table~\ref{tab:related-work-comparison} shows a comparison table summarizing key related works on cattle lameness detection, highlighting aspects such as data source, method type, model, sample size, and reported accuracy.

\begin{table}[hbt!]
\centering
\small
\caption{Comparison of Related Work on Cattle Lameness Detection}
\label{tab:related-work-comparison}
\begin{tabular}{@{}p{1cm}p{3.2cm}p{3.8cm}p{2cm}p{1.5cm}@{}}
\toprule
\textbf{Ref.} & \textbf{Data Source} & \textbf{Model Type} & \textbf{Sample Size} & \textbf{Accuracy (\%)} \\
\midrule
\cite{VanHertem2014} & 3D cameras in barns & Logistic Regression with extracted features & $\sim$1700 frames & 81.2 \\
\cite{Wu2020} & Barn videos, object detection via YOLO & LSTM with leg motion features & 5000 frames & 98.6 \\
\cite{Qiao2022} & Controlled video environment & C3D + ConvLSTM hybrid model & 15 cows & 90.3 \\
\cite{Barney2023} & Multi-cattle surveillance video & Modified Mask-RCNN for pose tracking & Not reported & -- \\
\cite{Russello2024} & Farm video data & Pose estimation + locomotion traits & -- & 80.1 \\
\cite{McDonagh2021} & Pre-calving cow monitoring & CNN-based behavior classifier & 46 cows & 83.0 \\
\textbf{This study} & YouTube videos (diverse environments) & End-to-end 3D CNN / ConvLSTM2D & 50 videos & 90.0 / 85.0 \\
\bottomrule
\end{tabular}
\end{table}

\section{Materials and Methods}
\label{sec2}

\begin{figure}[hbt!]
\centering
\includegraphics[width=0.95\linewidth]{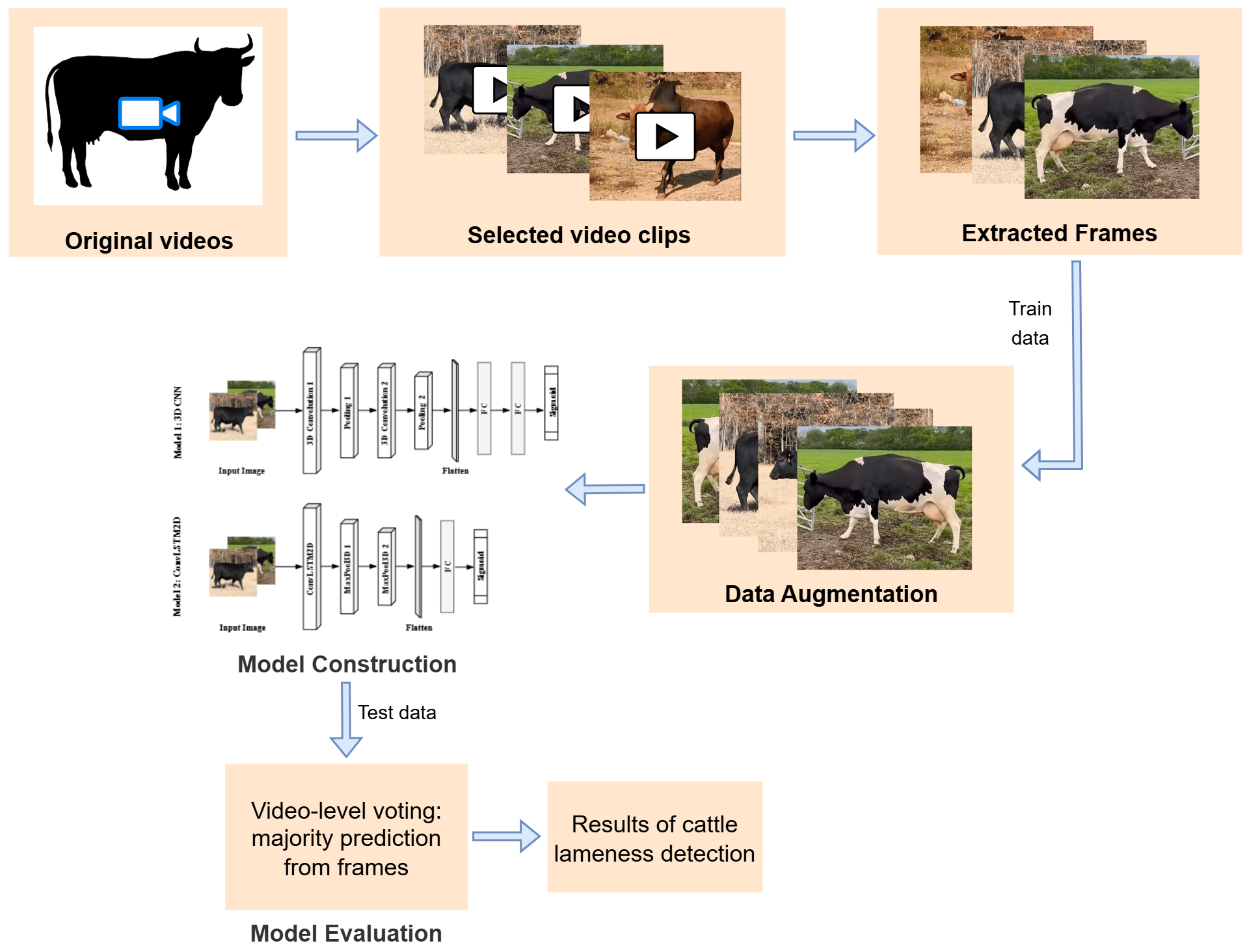}
\caption{Overall workflow of the cattle lameness classification model.}
\label{img01}
\end{figure}

Figure \ref{img01} illustrates the deep learning-based cattle lameness detection pipeline. The process begins with video data, from which frames are extracted during the data preprocessing stage. These extracted frames undergo data augmentation to enhance model generalization, producing a diverse training dataset. The model construction phase involves training two deep learning architectures: 3D CNN and ConvLSTM2D, denoted as $f_{\text{CNN3D}}(X)$ and $f_{\text{ConvLSTM2D}}(X)$, respectively. Finally, the models are evaluated using test data in the cattle classification stage to assess their performance.

\subsection{Dataset}

The data used in previous relevant research papers were collected following a similar approach. First, a specific cattle farm, including cows, was selected. Then, different types of video cameras were installed in convenient locations on the grid. The cameras captured videos of the cattle from desired angles (e.g., walking, eating, and moving) for a specific amount of time. These videos were processed, and only the relevant parts were used for further analysis. Additionally, many studies performed locomotion scoring (a manual lameness detection approach) to classify cattle as lame or non-lame. Based on this classification, unique identification was assigned to the cattle, and tracking of both classes was maintained. Later, deep learning models were trained and assessed using the data.

\subsubsection{Data collection}

The data collection process for this study was quite different from that of previous studies. We used an online platform to collect freely accessible cattle walking videos, specifically from YouTube. Publicly available data are commonly used in many research fields; for example, one of our previous articles \cite{Sohan2023} demonstrated that YouTube videos are frequently used in deepfake video detection research. In order to create an inclusive classification model, we created a diverse dataset. Therefore, the following features were incorporated with the video: there were different cattle in different environments, including intensive, semi-intensive, and extensive cattle farming; the cattle walked in different directions, including left to right and right to left; and the videos were captured from various views, including front, side, and back. 

\begin{figure}[!t]
    \centering
    \resizebox{0.98\textwidth}{!}{%
        \begin{minipage}{\textwidth}
            \centering
            \begin{minipage}{0.327\textwidth}
                \includegraphics[width=\linewidth]{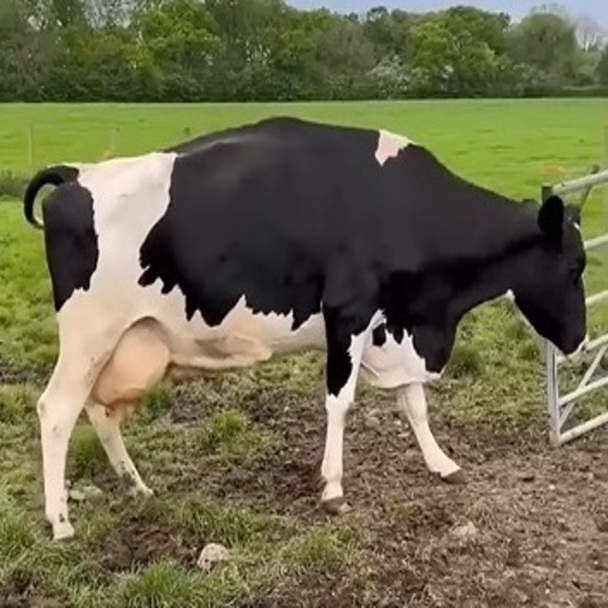}
            \end{minipage}
            \hfill
            \begin{minipage}{0.327\textwidth}
                \includegraphics[width=\linewidth]{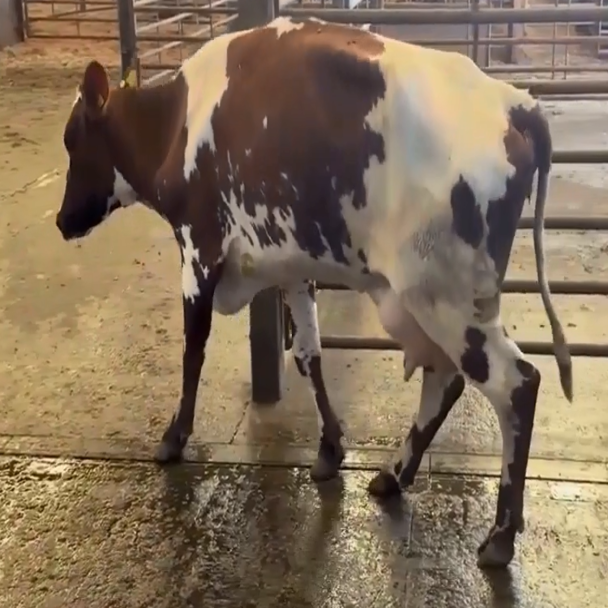}
            \end{minipage}
            \hfill
            \begin{minipage}{0.327\textwidth}
                \includegraphics[width=\linewidth]{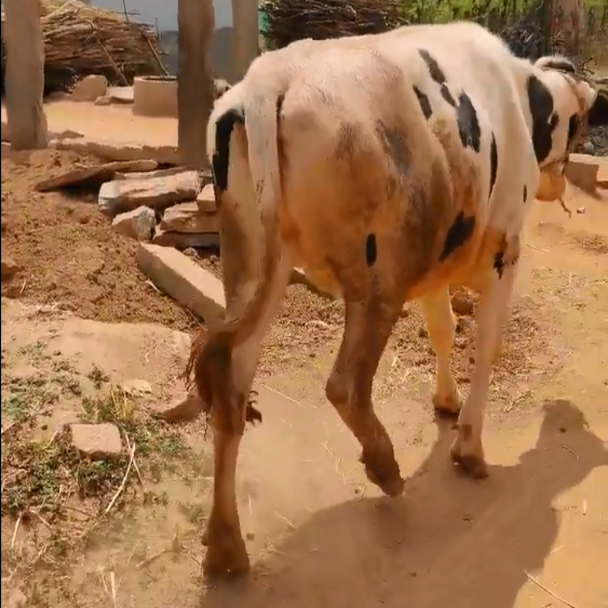}
            \end{minipage}

            \vspace{0.3cm}

            \begin{minipage}{0.327\textwidth}
                \includegraphics[width=\linewidth]{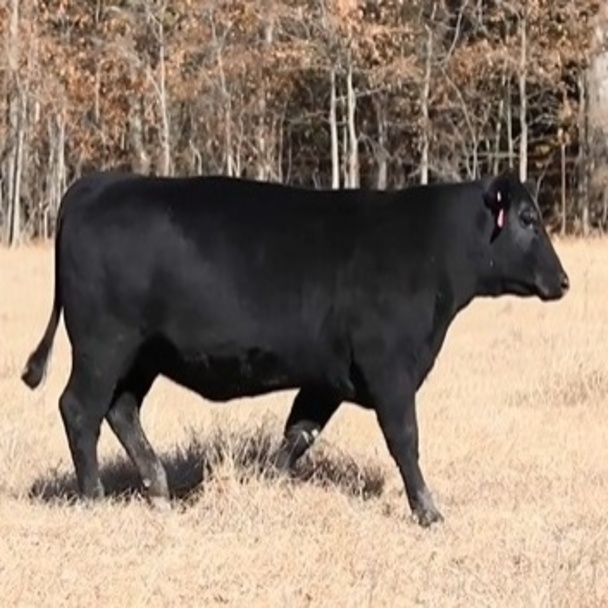}
            \end{minipage}
            \hfill
            \begin{minipage}{0.327\textwidth}
                \includegraphics[width=\linewidth]{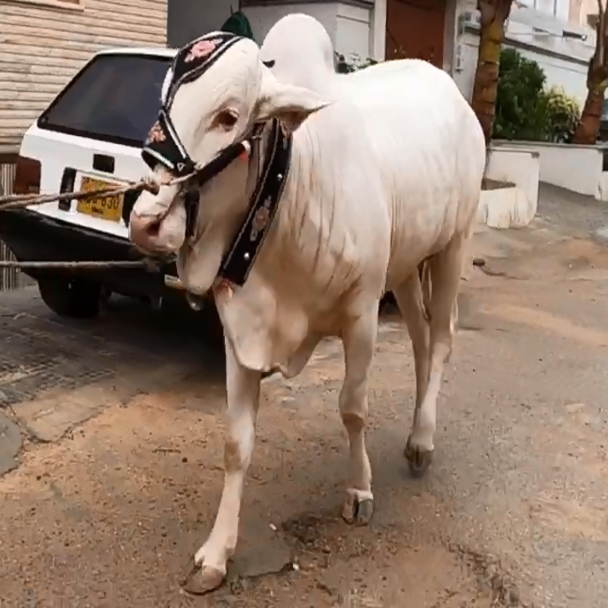}
            \end{minipage}
            \hfill
            \begin{minipage}{0.327\textwidth}
                \includegraphics[width=\linewidth]{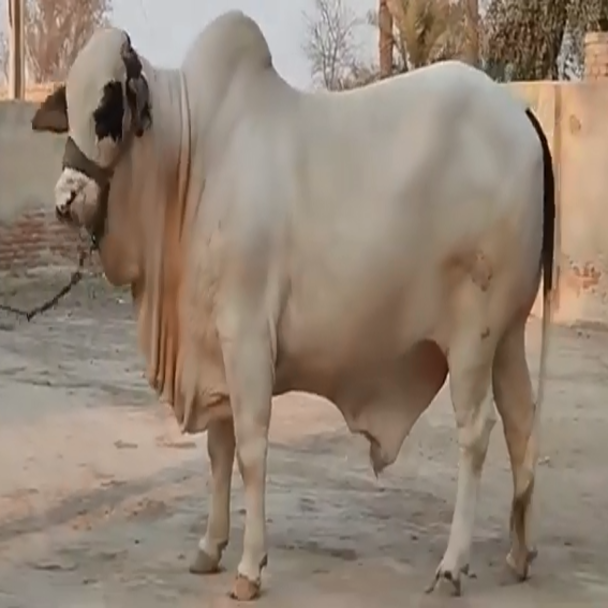}
            \end{minipage}
        \end{minipage}%
    }
    \caption{Example frames collected from video sequences of cattle used in this study. The upper set depicts frames from lame cattle, whereas the lower set illustrates frames from normal cattle.}
    \label{fig:all_images}
\end{figure}

\subsubsection{Basic Information}

\begin{table}[hbt!]
\caption{Summary of the dataset statistics.}
\label{tab:dataset-summary}
\centering
\begin{tabular}{lccc}
\toprule
\textbf{Class} & \textbf{Video Count} & \textbf{Duration (s)} & \textbf{Frame Count} \\
\midrule
Normal & 25 & 134.46 & 4050 \\
Lame & 25 & 148.91 & 5900 \\
\midrule
Total & 50 & 283.37 & 9950 \\
\bottomrule
\end{tabular}
\end{table}

To create the dataset, a total of 50 videos featuring 42 cattle were considered, with an average duration of 5.67 seconds per video. We ensured an equal class distribution of the data, meaning that 25 lame and 25 nomal cattle videos were included. Table~\ref{tab:dataset-summary} presents the dataset distribution, showing the number of videos, total duration in seconds, and the corresponding frame counts for the normal and lame classes. The normal class videos have a cumulative duration of 134.46 seconds and consist of 4050 frames, whereas the lame class videos have a total duration of 148.91 seconds and comprise 5900 frames. Overall, the dataset includes 50 videos with a combined duration of 283.37 seconds and a total of 9950 frames. Besides, Figure \ref{fig:all_images} displays example frames from cattle video sequences, with upper three images representing lame cattle and the other representing normal cattle. The original videos had varying resolutions due to collection from different online sources. To standardize the input data, all videos were resized to a resolution of \(500 \times 500\) pixels. 

\subsection{Data used in this study}

\begin{table}[hbt!]
\centering
\caption{Basic information of the dataset used in this study compared with the most relevant datasets in the literature}
\label{tab1}
\begin{tabular}{p{1.5cm}p{0.9cm}p{0.9cm}p{4.5cm}p{3cm}}
\toprule
\textbf{Study} & \textbf{Train Frames} & \textbf{Test Frames} & \textbf{Test Environment} & \textbf{Focused Activity} \\
\midrule
\cite{tun2024} & 1376 & 345 & Within the barn & Intensive farming \\
\cite{Kang2020} & 1000 & 500 & Passing alley & Intensive farming \\
\cite{Wu2020} & 490 & 210 & Walked along a narrow path & Intensive farming \\
\cite{Myint2024} & 690 & 341 & Barn to milking parlor pathway & Intensive farming \\
\cite{Jia2023} & 600 & 120 & Corridor after the milking parlor & Intensive farming \\
This study & 1500 & 500 & Indoor and outdoor environment & Semi-intensive farming \\
\bottomrule
\end{tabular}
\end{table}

Table \ref{tab1} provides an overview of the data used in this research alongside datasets from previous cattle lameness detection studies. The table compares these studies based on the number of training and testing frames, the test environment, and the focused activity. As previously mentioned, the dataset consists of a total of 50 videos, including 25 normal and 25 lame cattle videos. Out of these, 30 videos (15 normal and 15 lame) were used for training the model, while the remaining 20 videos (10 normal and 10 lame) were used for testing. Previous studies focused on intensive farming environments, where cattle were primarily managed in indoor grazing conditions and data was collected from specific, limited locations such as within a barn, a passing alley, a narrow path, and corridors related to milking parlors. In contrast, this study focuses on a semi-intensive farming setting and incorporates both indoor and outdoor environments. This approach captures a wider range of cattle movement behaviors, lighting conditions, and surface variations, potentially leading to a more generalizable model for cattle lameness detection.

\subsubsection{Data pre-processing}

Preprocessing in this investigation was conducted in several distinct steps, including frame extraction and data augmentation. The overall average frame rate was approximately 35.12 frames per second. However, due to the high volume of frames, only 25 images were randomly selected from each training video, resulting in a total of 750 images. These images were then cropped to a 224×224 pixel ratio. However, data augmentation is a widely recognized method for addressing limited data issues and helps to create robust and sustainable classification models. Numerous data augmentation techniques are available, as demonstrated by Jia et al. \cite{Jia2023}, who employed nine different methods, such as Mixup, Cutout, CutMix, etc. 

The primary goal of augmentation in this study is to increase the learning capability of the training model in diverse ways. This study considered only image flipping as an augmentation technique, applied solely to the training data. his decision was driven by several factors, such as:

\begin{itemize} 
    \item  Horizontal flipping realistically simulates left-to-right and right-to-left walking, which are common in cattle locomotion videos. This increases the model's ability to generalize across different movement orientations without altering the fundamental gait characteristics.
    \item Vertical flipping or rotation was also avoided because it inverts the image, resulting in an anatomically impossible scenario with the cattle's head at the bottom and hooves at the top of the frame. This can hinder the model’s ability to learn realistic gait patterns.
    \item The dataset used in this study already includes a variety of backgrounds, as the videos were collected from different online sources. This is why augmentation techniques that artificially alter the background, such as random background replacement, were considered unnecessary and could even introduce misleading variability. Similarly, crop jitter was not utilized due to the risk of removing important contextual information from the video frames.
\end{itemize}

Therefore, to maintain the integrity of the gait features and avoid introducing misleading transformations, horizontal flipping was considered the most suitable augmentation method for this study. In this continuity, all 30 training videos (750 frames) were horizontally flipped, generating a total of 1500 images. In test data, 500 images were extracted from 20 videos. However, no data augmentation was performed on the test dataset to preserve integrity and ensure a fair assessment of the model's generalization ability.

\subsection{Model Construction}

\begin{algorithm}
\caption{Pseudo-code of the Proposed Classification Model}\label{Alg1}
\begin{algorithmic}[1]
\Require 
\begin{tabular}[t]{@{}l@{}}
\textbullet \ Training dataset $\mathcal{D}_{\text{train}}$ and testing dataset $\mathcal{D}_{\text{test}}$,
where $\mathcal{D} = \{(V_i, l_i)\}$,\\ \hspace{3mm}$V_i$ is the $i$-th video, $l_i \in \{\text{normal, lame}\}$ \\
\textbullet \ Number of frames per video $N_{\text{frames}}$ \\
\textbullet \ Target size for frame resizing $S_{\text{target}}$ \\
\textbullet \ Data augmentation probability $p_{\text{aug}}$ \\
\textbullet \ Batch size $B$, Number of epochs $E$
\end{tabular}
\Ensure Trained 3DCNN or ConvLSTM2D model and performance metrics

\State \textbf{Data Preparation}
\ForAll{$(V_i, l_i) \in \mathcal{D}_{\text{train}} \cup \mathcal{D}_{\text{test}}$}
    \State $F_i \Leftarrow \text{Pad/Truncate}\big(\text{ResizeFrames}(\text{ExtractFrames}(V_i, \text{max\_frames}=25), S_{\text{target}}), N_{\text{frames}}\big)$
    \If{$(V_i, l_i) \in \mathcal{D}_{\text{train}}$}
        \State $(X, y) \Leftarrow (X \cup \{F_i\}, y \cup \{l_i\})$
    \Else
        \State $(X_{\text{test}}, y_{\text{test}}) \Leftarrow (X_{\text{test}} \cup \{F_i\}, y_{\text{test}} \cup \{l_i\})$
    \EndIf
\EndFor
\State Apply horizontal flip augmentation to $X_{\text{train}}$ with probability $p_{\text{aug}}$
\State Normalize $X_{\text{train}}, X_{\text{test}}$ and one-hot encode labels

\State \textbf{Model Training}
\State $M \Leftarrow \text{CreateModel}(N_{\text{frames}}, S_{\text{target}})$
\State $M \Leftarrow \text{TrainModel}(M, X_{\text{train}}, y_{\text{train}}, B, E)$

\State \textbf{Model Evaluation}
\State $\hat{y} \Leftarrow M(X_{\text{test}})$
\State $(\text{metrics}, \text{confusion matrix}) \Leftarrow \text{Evaluate}(y_{\text{test}}, \hat{y})$
\State \Return $M$, metrics, confusion matrix
\end{algorithmic}
\end{algorithm}

Algorithm \ref{Alg1} outlines the pseudocode of the classfication model. It begins with data preparation by processing the training and testing datasets ($\mathcal{D}_{\text{train}}$, $\mathcal{D}_{\text{test}}$), extracting frames from each video, resizing them to a target size ($S_{\text{target}}$), and ensuring each video has $N_{\text{frames}}$ frames through padding or truncation, creating input data ($X$) and labels ($y$). Training data is augmented with random horizontal flipping (probability $p_{\text{aug}}$). Pixel values are normalized, and behavior labels (``normal" and ``lame") are encoded numerically. Next, model training involves creating a 3D CNN or ConvLSTM2D model, configured using $N_{\text{frames}}$ and $S_{\text{target}}$, and training it on the prepared data ($X_{\text{train}}$, $y_{\text{train}}$) using a batch size ($B$) and epochs ($E$). The model learns patterns indicative of different behavior classes. Finally, model evaluation uses the trained 3D CNN model to predict behavior labels ($\hat{y}$) for the testing dataset ($X_{\text{test}}$), comparing them to true labels ($y_{\text{test}}$) to assess performance. Metrics like accuracy are calculated, and a confusion matrix is generated. The trained model, performance metrics, and confusion matrix are returned as output.

\subsubsection{3D CNN}

\begin{figure}[hbt!]
\centering\includegraphics[width=0.80\linewidth]{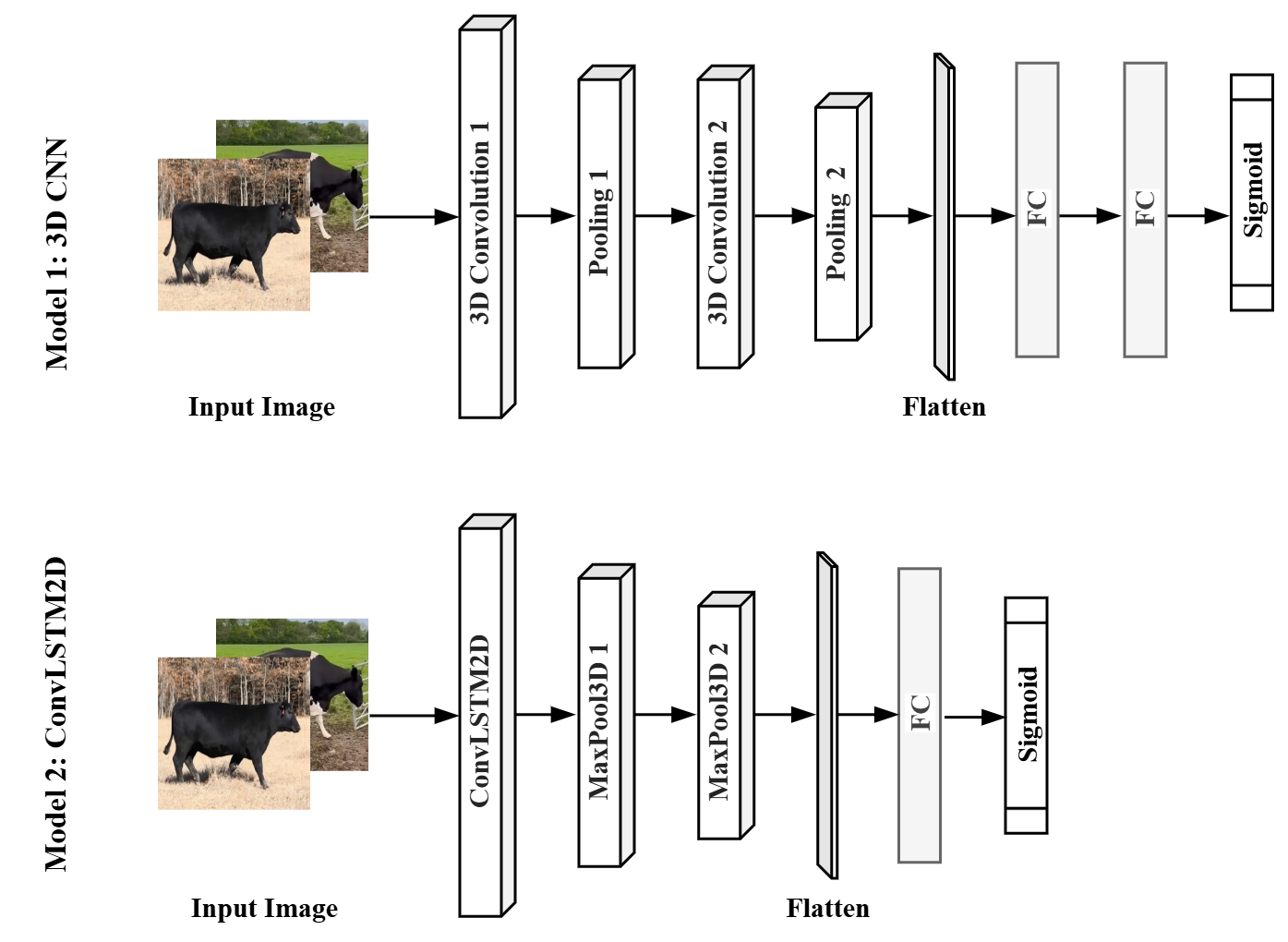}
\caption{Architectures of the 3D CNN and ConvLSTM2D models used for the cattle classification.}
\label{fig:model_comparison_updated}
\end{figure}

CNNs are widely used deep learning models for image processing tasks. 3D CNNs are an extension of these and can handle 3D data containing spatiotemporal information, such as videos. They utilize 3D convolutional kernels, which enable them to focus on spatial features extracted from input data and distinguish themselves from traditional 2D CNNs. Because of their ability to learn both spatial and temporal features, this study employed a 3D CNN model using the Keras Sequential API to perform binary classification, where movement patterns are assessed over time. The network accepts input tensors of shape $(25, 224, 224, 3)$, where 25 denotes the number of consecutive frames, $224 \times 224$ represents the spatial resolution of each frame, and 3 corresponds to the RGB color channels. 

As the Figure \ref{fig:model_comparison_updated} shows, the model architecture begins with a 3D convolutional layer comprising 32 filters with a kernel size of $(3 \times 3 \times 3)$, employing a ReLU activation function and \texttt{same} padding to preserve the original spatial and temporal dimensions. This is followed by a 3D max pooling layer with a pooling size of $(2 \times 2 \times 2)$ to reduce dimensionality and computational load. A second convolutional layer with 64 filters of size $(3 \times 3 \times 3)$ and ReLU activation is subsequently applied, again followed by a $(2 \times 2 \times 2)$ max pooling operation. The extracted feature maps are then flattened into a one-dimensional vector, which is passed through two fully connected layers with 128 and 64 neurons, respectively, each utilizing a ReLU activation function. To reduce the risk of overfitting, dropout layers with a dropout rate of 0.5 are inserted after each dense layer. The final output layer consists of a single neuron with a sigmoid activation function, yielding a probability score for binary classification. The model is compiled using the Adam optimizer, binary cross-entropy as the loss function, and accuracy as the evaluation metric.

\subsubsection{ConvLSTM2D}

ConvLSTM2D is a deep learning architecture that integrates the strengths of spatial feature extraction and temporal dependency modeling through the combination of CNNs and LSTM networks, respectively. This makes it particularly effective for understanding patterns and movements in sequential data. Followed by the previous model, the input layer is similar, processing video frames in the same format. In Figure \ref{fig:model_comparison_updated}, model 2 employed in this study begins with a ConvLSTM2D layer. Following the ConvLSTM2D layer, two successive 3D max pooling layers with pooling sizes of $(1 \times 2 \times 2)$ are applied to downsample the spatial dimensions while maintaining the temporal dimension. The output is then flattened into a one-dimensional feature vector. A fully connected dense layer with 128 neurons and ReLU activation is employed for high-level feature abstraction, followed by a dropout layer with a rate of 0.25 to reduce overfitting. Finally, a dense layer with sigmoid activation is used to classify cattle into ‘normal’ and ‘lame’ categories. The compilation process also follows the same procedure as in the previous model.

\subsection{Model Evaluation}

The four performance evaluation measures (accuracy, precision, recall, and F1-score) used in this study are associated with the confusion matrix; therefore, both are presented gradually. The values of all evaluation criteria range from 0\% to 100\%, where a higher value indicates stronger performance. In the mathematical framework (equation 1, 2, 3, and 4) for model evaluation,  $n$ represents the total instances, $y_i$ denotes the true label for instance $i$, and $\hat{y}_i$ signifies the model's predicted label. The indicator function $I(\text{condition})$ returns 1 if the condition is true, otherwise 0. Summation, denoted by $\sum_{i=1}^{n}$, calculates the cumulative value across all instances. These components enable quantitative assessment by comparing predicted ($\hat{y}_i$) and true ($y_i$) labels, facilitating calculation of the evaluation metrics, providing a comprehensive evaluation of model performance.

    \begin{equation}\label{e1}
    Accuracy = \frac{1}{n} \sum_{i=1}^{n} I(y_i = \hat{y}_i)
    \end{equation}

    \begin{equation}\label{e2}
    Precision = \frac{\sum_{i=1}^{n} I(y_i = 1 \land \hat{y}_i = 1)}{\sum_{i=1}^{n} I(\hat{y}_i = 1)}
    \end{equation}

    \begin{equation}\label{e3}
    Recall = \frac{\sum_{i=1}^{n} I(y_i = 1 \land \hat{y}_i = 1)}{\sum_{i=1}^{n} I(y_i = 1)}
    \end{equation}

    \begin{equation}\label{e4}
    F1-Score = \frac{2 \times Precision \times Recall}{Precision + Recall}
    \end{equation}

\section{Results}
\label{sec3}

This project was developed using the Python programming language and implemented within the Google Colab Pro environment. Data preprocessing tasks were handled by OpenCV and MoviePy, supplemented by Keras' ImageDataGenerator for data augmentation. Furthermore, TensorFlow, NumPy, Pandas, Scikit-learn, and Matplotlib were employed for model construction, training, performance evaluation, and visualization.

\begin{figure}[hbt!]
    \centering
    \subfloat[3DCNN]{\includegraphics[width=0.48\linewidth]{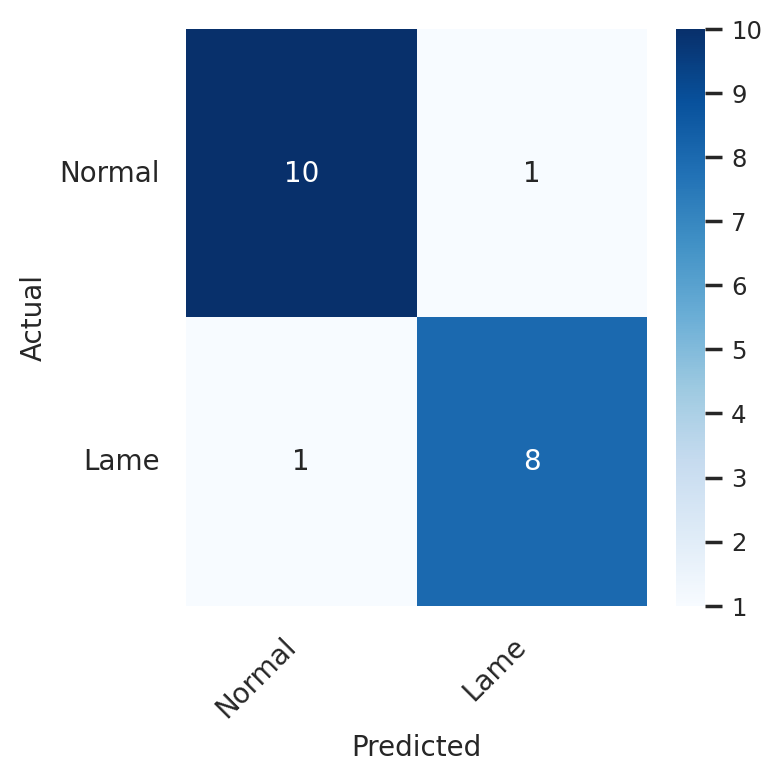}%
    \label{fig:subfig1}}
    \hfil
    \subfloat[ConvLSTM2D]{\includegraphics[width=0.48\linewidth]{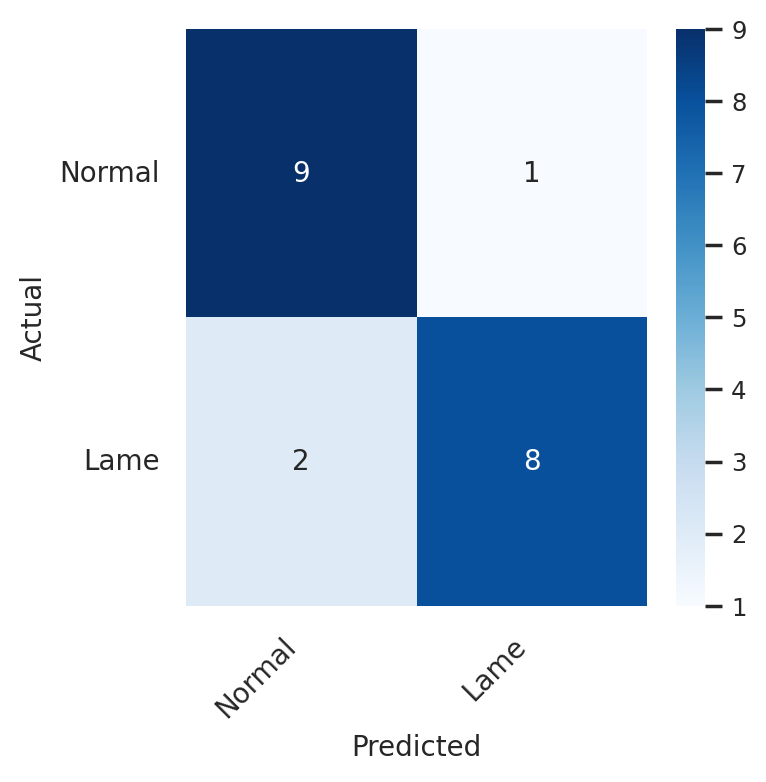}%
    \label{fig:subfig2}}
    \caption{Confusion matrix of both models under the same test data.}
    \label{fig2}
\end{figure}

The dataset utilized in this study comprised 50 videos, with 50\% depicting defective cattle and 50\% depicting non-defective cattle. Of these 50 videos, 30 were used for model training, and 20 were used for validating the 3D CNN and ConvLSTM2D models. In the test phase, 500 frames were selected from 20 test videos (10 normal and 10 lame) to evaluate the performance of both models individually. Frame sampling (25 frames per video) and frame size ($224 \times 224$ pixels) were kept consistent with the training phase to ensure a fair assessment. Augmentation techniques were deliberately avoided for the test data to preserve the original video characteristics, enabling an unbiased evaluation on unaltered, real-world data. Each video was classified based on a majority vote from its 25 frame-level predictions; a video was considered correctly classified if the model accurately predicted the majority of its frames. Ultimately, the classification report was presented based on video-level performance. This method ensures a robust evaluation while accommodating the variable lengths and conditions present in real-world video data. The classification results achieved by the models were calculated and presented here.

\begin{table}[hbt!]
\centering
\caption{Lameness classification results of proposed 3D CNN and ConvLSTM2D}
\label{tab2}
\small
\begin{tabular}{lcccc}
\toprule
\textbf{Model} & \textbf{Accuracy} & \textbf{Precision} & \textbf{Recall} & \textbf{F1-score} \\
\midrule
3DCNN & 90 & 90.9 & 90.9 & 90.91 \\
ConvLSTM2D & 85 & 90 & 81.82 & 85.71 \\
\bottomrule
\end{tabular}
\end{table}

The confusion matrices for both models are presented in Figure \ref{fig2}. As shown in Figures \ref{fig:subfig1} and \ref{fig:subfig2}, a small number of misclassifications were observed, including 1 false positives and 1 false negative for 3D CNN, and 1 false positives and 2 false negatives for ConvLSTM2D. The 3D CNN model demonstrated notable performance in classifying the cattle data. Table \ref{tab2} indicates an accuracy of 90\%, along with strong average precision, recall, and F1-scores. On the other hand, ConvLSTM2D achieved comparatively lower performance, with accuracy, precision, recall, and F1-score values of 85\%, 90\%, 81.82\%, and 85.71\%, respectively. Overall, the confusion matrices and classification reports highlight the models’ effectiveness in classifying healthy and defective cattle.

\subsection{Qualitative Evaluation of Model Predictions}

\begin{figure}[hbt!]
    \centering
    \resizebox{0.99\textwidth}{!}{%
        \begin{minipage}{\textwidth}
            \centering
            \begin{minipage}{0.24\textwidth}
                \includegraphics[width=\linewidth]{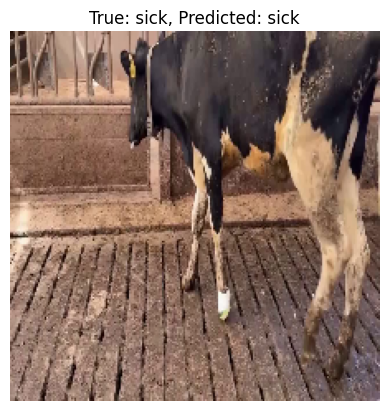}
            \end{minipage}
            \hfill
            \begin{minipage}{0.24\textwidth}
                \includegraphics[width=\linewidth]{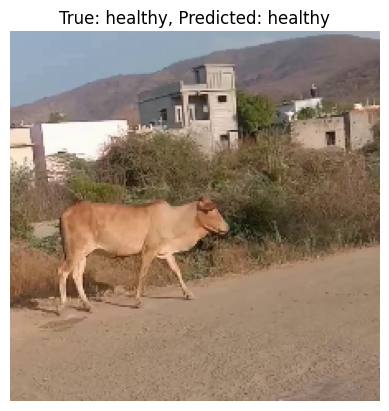}
            \end{minipage}
            \hfill
            \begin{minipage}{0.24\textwidth}
                \includegraphics[width=\linewidth]{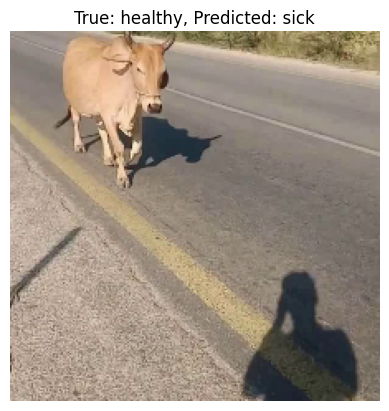}
            \end{minipage}
            \begin{minipage}{0.24\textwidth}
                \includegraphics[width=\linewidth]{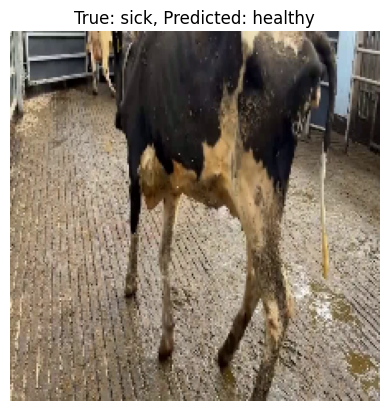}
            \end{minipage}

            \vspace{0.3cm}

            \begin{minipage}{0.24\textwidth}
                \includegraphics[width=\linewidth]{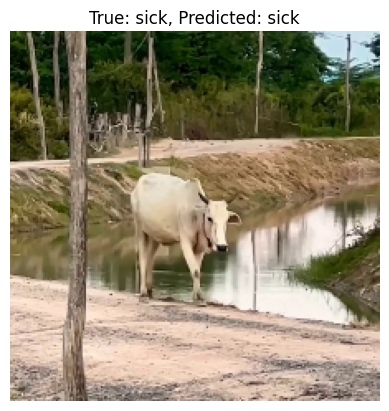}
            \end{minipage}
            \hfill
            \begin{minipage}{0.24\textwidth}
                \includegraphics[width=\linewidth]{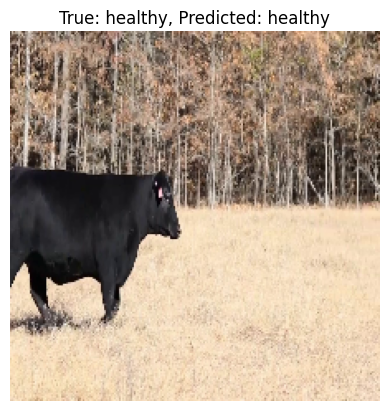}
            \end{minipage}
            \hfill
            \begin{minipage}{0.24\textwidth}
                \includegraphics[width=\linewidth]{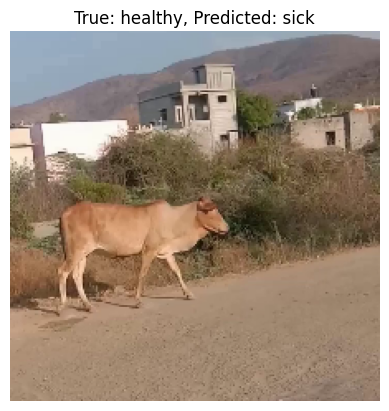}
            \end{minipage}
            \begin{minipage}{0.24\textwidth}
                \includegraphics[width=\linewidth]{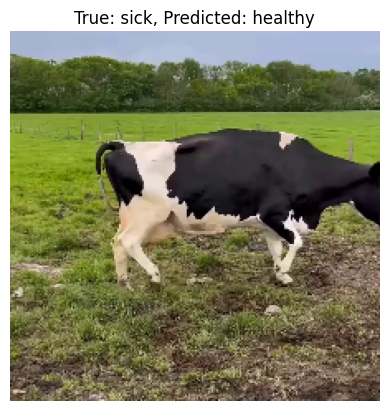}
            \end{minipage}
        \end{minipage}%
    }
    \caption{Sample frames from the test dataset illustrating classification outcomes of the two models. The upper row displays results from the 3D CNN model, and the lower row shows results from the ConvLSTM2D model. Each row includes examples of true positive (TP), true negative (TN), false positive (FP), and false negative (FN) predictions, respectively. These frames qualitatively highlight both correct classifications and misclassification scenarios.}
    \label{fig:qualitative}
\end{figure}

Along with the numerical evaluation, we also reviewed some of the test videos to understand where the models performed well or struggled. This qualitative analysis helps highlight practical strengths and weaknesses that may help to understand the model performance more clearly. Figure \ref{fig:qualitative} presents representative frames from the test set illustrating the classification outcomes of both models. The top row corresponds to the 3D CNN model, while the bottom row shows the results of the ConvLSTM2D model. Each row includes one sample each of true positive (TP), true negative (TN), false positive (FP), and false negative (FN) predictions.

\begin{enumerate}
    \item \textbf{Correct Classifications:} The 3D CNN model consistently produced correct predictions when gait abnormalities were visually prominent, for instance, limping or dragging a leg (TP, first column of Figure \ref{fig:qualitative}). Similarly, the second column shows a healthy cow walking with a balanced gait and straight backline, which the 3D CNN correctly classified as a TN. Notably, the same frame was misclassified by ConvLSTM2D, which labeled the healthy cow as lame. This inconsistency may indicate a limitation of the ConvLSTM2D model in handling slight movements in walking or staying reliable when appearances change.
    \item \textbf{Misclassifications:} Though the overall results show strong performance, both models exhibited some misclassifications. The third and fourth columns of Figure \ref{fig:qualitative} illustrate misclassification cases. The 3D CNN produced a FP in one instance, likely due to motion blur or transient posture changes.
    \item \textbf{Postural Differences in Lame vs. Healthy Cattle:} In addition to prediction behavior, we observed distinct physical posture differences between lame and healthy cattle. As shown in Figure \ref{fig:all_images}, the lame cow’s body forms a noticeable arc from the head to the tail. In this frames, the typical shoulder hump appears less defined, almost blending into the backline, possibly as a way to relieve pain or pressure while walking or standing. On the other side, the figure shows healthy cows are with a straight and level backline, an upright head, and a sharp, clearly visible shoulder hump. This posture indicates a normal body mechanics and balanced movement. These visual distinction offers meaningful insights into lameness and could support future model development by combining geometric body features with spatiotemporal motion data.
\end{enumerate}

Overall, the qualitative examples in Figure \ref{fig:qualitative} align with the trends seen in the confusion matrices (Figure \ref{fig2}) and classification scores (Table \ref{tab2}), and they further emphasize the ConvLSTM2D model’s relative weakness in cases with subtle or ambiguous visual patterns.

\section{Discussion}
\label{sec4}

Based on available information, all previous research on cattle lameness, sickness, or gait abnormality detection using computer vision technology has been conducted with very specific and limited data types. Typically, these studies focused on a single cattle barn, where video cameras were set up to monitor the cattle. The collected video data were then used for further analysis, including object detection, pose estimation, feature extraction, and classification. In contrast, this study utilized cattle videos from online platforms, which encompass multi-dimensional features. This departure from previous studies allowed for the development of a more universal and comprehensive lameness detection model that is suitable for large-scale cattle farms. The results of this study also validate the applicability of the classification model.

\begin{table}[bpt!]
\centering
\caption{Performance comparison of this study with existing cattle lameness classification models}
\label{tab3}
\begin{tabular}{p{1.5cm}p{2.8cm}p{1.5cm}p{1.5cm}p{1.5cm}p{1.5cm}}
\toprule
\textbf{Study} & \textbf{Classifier} & \textbf{Accuracy} & \textbf{Precision} & \textbf{Recall} & \textbf{F1-score} \\
\midrule
\cite{VanHertem2014} & Logistic Regression & 81.18 & 75 & 47.06 & 57.83 \\
\cite{Qiao2022} & C3D-ConvLSTM & 90.32 & 90.62 & 88.76 & 89.68 \\
\cite{Russello2024} & SVM & 80.07 & -- & -- & 78.70 \\
\cite{Myint2024} & AdaBoost & 77.9 & -- & -- & -- \\
\cite{zhao2018} & DT & 90.18 & -- & -- & -- \\
\cite{jia2025} & CNN & 90.21 & -- & -- & -- \\
\cite{myint2024cattle} & SVM & 83.58 & 86.04 & 88.43 & 87.21 \\
\cite{tun2024} & KNN & 73.33 & 89.08 & 75.28 & 81.60 \\
This study & 3DCNN & 90 & 90.9 & 90.9 & 90.91 \\
\bottomrule
\end{tabular}
\begin{tablenotes}
\small
\item \textit{-- No specific data was found}
\end{tablenotes}
\end{table}

Table \ref{tab3} presents a performance comparison between eight prior studies and the proposed model. The table evaluates models based on accuracy, precision, recall, and f1-score. The final row of the table displays the results of this study, which employed a 3D CNN model, while the preceding rows summarize the findings from other relevant studies that utilized various machine learning and deep learning classifiers. The model used in this study achieved a notable accuracy of 90\%. More significantly, it demonstrated the highest precision among the studies where precision values were available, reaching 90.9\%. This indicates that the model exhibits a strong ability to minimize false positives, correctly identifying positive cases with high confidence. Additionally, the 3D CNN model also yielded a robust F1-score of 90.91\%. When compared to other approaches, the 3D CNN model in this study demonstrated competitive performance, whereas study \cite{Qiao2022} employed C3D-ConvLSTM achieved a slightly higher accuracy of 90.32\% but a lower precision of 90.62\%.

The results indicate that deep learning models generally outperform traditional machine learning approaches, such as SVM, Logistic Regression, and KNN. Furthermore, the findings of this study demonstrate the effectiveness of 3D CNNs for the video classification, as they efficiently capture both spatial and temporal information and offer well-balanced performance across multiple evaluation metrics.

This project's computational aspects were implemented in Python using TensorFlow and Keras, and conducted in the Google Colab Pro environment equipped with an NVIDIA Tesla T4 GPU. The training set comprised 1,500 augmented frames from 30 videos, while the evaluation was performed on 500 frames from 20 test videos. The 3D CNN model demonstrated faster convergence and lower computational demand, with each training epoch requiring approximately 1 minutes. In contrast, the ConvLSTM2D model, which incorporates recurrent layers for temporal modeling, incurred slightly higher computational overhead, requiring about 1-2 minutes per epoch. Despite these differences, both models remained computationally feasible to train and evaluate using a cloud-based consumer-grade platform.

\subsection{Limitation and future plan}

While this study demonstrates the potential of deep learning for cattle lameness detection, certain limitations may affect its broader applicability and real-world deployment, potentially impacting the generalizability of the findings to broader contexts. The following points outline these limitations:

\begin{itemize}
    \item All of the videos were sourced from YouTube. The cattle abnormality was determined based on the video titles and descriptions; however, it was not possible to individually verify the class and level of abnormality.
    \item Although the simplicity of the proposed methodology (which directly employs classification algorithms) enables successful implementation with the video data, it is also necessary to experiment with intermediate processing steps such as object detection and pose estimation.
    \item While this study utilized a greater number of training and testing frames compared to many similar studies, the total number of video samples remains relatively limited, which may affect the model's generalizability and robustness.
    \item We did not benchmark recent transformer-based video models (e.g., TimeSformer, Video-Swin). Future work will explore.
\end{itemize}

To address the limitations outlined in this study and further enhance the applicability of our approach, future work should focus on expanding the dataset with a greater variety of video sources, including direct recordings from cattle farms, which would ensure a more robust representation of real-world scenarios, encompassing diverse cattle breeds, environmental conditions, and lameness severities, as well as addressing potential biases in online video content. In addition, while this study initially utilizes a direct classifier, future research will explore the impact of feature engineering through object detection, pose estimation, and feature extraction to assess the dataset and model performance. Overall, this study is an initial initiative of a benchmark cattle lameness detection dataset, future work will be introducing a complete benchmark of the approach.

\section{Conclusions}
\label{sec5}

This paper presents a cattle lameness video dataset and deep learning-based methods for effectively distinguishing between lame and non-lame cattle from video data. Compared to existing datasets used for cattle lameness detection, this study collected data entirely from online sources, providing more comprehensive information across diverse environments, scenes, and activities. Additionally, two deep learning models, 3D CNN and ConvLSTM2D, were used to classify normal and lame cattle. From the dataset, 1,500 images were extracted for training the models, while 500 images were used for validation. The experimental results demonstrated improved or at least comparable performance to existing cattle lameness classification models. In conclusion, this study serves as a reference for detecting cattle lameness or behavior without relying on traditional multi-step approaches.

\vspace{4mm}

\hspace{-5mm}\textbf{Author Contributions}: Md Fahimuzzman Sohan contributed to conceptualization, methodology, formal analysis, software implementation, and manuscript drafting. Raid Alzubi contributed to data curation, validation, and interpretation of results. Hadeel Alzoubi and Eid Albalawi contributed to methodology refinement and critical manuscript review. A. H. Abdul Hafez supervised the project, provided resources, and contributed to manuscript editing and final approval. All authors have read and approved the final manuscript.

\vspace{4mm}

\hspace{-5mm}\textbf{Data availability} The dataset used and analysed during the current study come from third parties that are publicly available. We commit to open-sourcing the codebase and datasets after the paper is accepted.

\section*{Declarations}

\textbf{Conflict of interest} The authors declare no competing interests.\\ 

\vspace{1mm}

\hspace{-5mm}\textbf{Ethics approval} This study used publicly available video recordings of dairy cows from YouTube to develop and evaluate a spatiotemporal deep learning model for lameness detection. No animals were directly involved in experiments or handling, and all videos were collected in accordance with the platform’s terms of service. Ethical approval was not required as the study did not involve any live animal experimentation.



\begin{thebibliography}{00}

\bibitem{Thomas2024survey}
Thomsen, P.T.; Shearer, J.K.; Houe, H. Prevalence of lameness in dairy cows: A literature review. \emph{Vet. J.} \textbf{2023}, \emph{295}, 105975. https://doi.org/10.1016/j.tvjl.2023.105975

\bibitem{Qiao2021}
Qiao, Y.; Kong, H.; Clark, C.; Lomax, S.; Su, D.; Eiffert, S.; Sukkarieh, S. Intelligent perception-based cattle lameness detection and behaviour recognition: A review. \emph{Animals} \textbf{2021}, \emph{11}, 3033. https://doi.org/10.3390/ani11113033

\bibitem{Kang2021}
Kang, X.; Zhang, X.D.; Liu, G. A review: Development of computer vision-based lameness detection for dairy cows and discussion of practical applications. \emph{Sensors} \textbf{2021}, \emph{21}, 753. https://doi.org/10.3390/s21030753

\bibitem{Jia2023}
Jia, Z.; Yang, X.; Wang, Z.; Yu, R.; Wang, R. Automatic lameness detection in dairy cows based on machine vision. \emph{Int. J. Agric. Biol. Eng.} \textbf{2023}, \emph{16}, 217--224. https://doi.org/10.25165/j.ijabe.20231605.8195

\bibitem{VanHertem2014}
Van Hertem, T.; Viazzi, S.; Steensels, M.; Maltz, E.; Antler, A.; Alchanatis, V.; Halachmi, I. Automatic lameness detection based on consecutive 3D-video recordings. \emph{Biosyst. Eng.} \textbf{2014}, \emph{119}, 108--116. https://doi.org/10.1016/j.biosystemseng.2013.12.006

\bibitem{Russello2024}
Russello, H.; van der Tol, R.; Holzhauer, M.; van Henten, E.J.; Kootstra, G. Video-based automatic lameness detection of dairy cows using pose estimation and multiple locomotion traits. \emph{Comput. Electron. Agric.} \textbf{2024}, \emph{223}, 109040. https://doi.org/10.1016/j.compag.2024.109040

\bibitem{Schlageter-Tello18}
Schlageter-Tello, A.; Van Hertem, T.; Bokkers, E.A.; Viazzi, S.; Bahr, C.; Lokhorst, K. Performance of human observers and an automatic 3D computer-vision-based locomotion scoring method to detect lameness and hoof lesions in dairy cows. \emph{J. Dairy Sci.} \textbf{2018}, \emph{101}, 6322--6335. https://doi.org/10.3168/jds.2017-13535

\bibitem{Jabbar2017}
Jabbar, K.A.; Hansen, M.F.; Smith, M.L.; Smith, L.N. Early and non-intrusive lameness detection in dairy cows using 3D video. \emph{Biosyst. Eng.} \textbf{2017}, \emph{153}, 63--69. https://doi.org/10.1016/j.biosystemseng.2016.11.012

\bibitem{McDonagh2021}
McDonagh, J.; Tzimiropoulos, G.; Slinger, K.R.; Huggett, Z.J.; Down, P.M.; Bell, M.J. Detecting dairy cow behavior using vision technology. \emph{Agriculture} \textbf{2021}, \emph{11}, 675. https://doi.org/10.3390/agriculture11070675

\bibitem{Kang2020}
Kang, X.; Zhang, X.D.; Liu, G. Accurate detection of lameness in dairy cattle with computer vision: A new and individualized detection strategy based on analysis of the supporting phase. \emph{J. Dairy Sci.} \textbf{2020}, \emph{103}, 10628--10638. https://doi.org/10.3168/jds.2020-18385

\bibitem{FernandezCarrion2020}
Fernández-Carrión, E.; Barasona, J.Á.; Sánchez, Á.; Jurado, C.; Cadenas-Fernández, E.; Sánchez-Vizcaíno, J.M. Computer vision applied to detect lethargy through animal motion monitoring: A trial on African swine fever in wild boar. \emph{Animals} \textbf{2020}, \emph{10}, 2241. https://doi.org/10.3390/ani10122241

\bibitem{Poursaberi2010}
Poursaberi, A.; Bahr, C.; Pluk, A.; Van Nuffel, A.; Berckmans, D. Real-time automatic lameness detection based on back posture extraction in dairy cattle: Shape analysis of cow with image processing techniques. \emph{Comput. Electron. Agric.} \textbf{2010}, \emph{74}, 110--119. https://doi.org/10.1016/j.compag.2010.07.003

\bibitem{Pluk2012}
Pluk, A.; Bahr, C.; Poursaberi, A.; Maertens, W.; Van Nuffel, A.; Berckmans, D. Automatic measurement of touch and release angles of the fetlock joint for lameness detection in dairy cattle using vision techniques. \emph{J. Dairy Sci.} \textbf{2012}, \emph{95}, 1738--1748. https://doi.org/10.3168/jds.2011-4711

\bibitem{Alsaaod2019}
Alsaaod, M.; Fadul, M.; Steiner, A. Automatic lameness detection in cattle. \emph{Vet. J.} \textbf{2019}, \emph{246}, 35--44. https://doi.org/10.1016/j.tvjl.2019.01.005

\bibitem{Fuentes2023}
Fuentes, A.; Han, S.; Nasir, M.F.; Park, J.; Yoon, S.; Park, D.S. Multiview monitoring of individual cattle behavior based on action recognition in closed barns using deep learning. \emph{Animals} \textbf{2023}, \emph{13}, 2020. https://doi.org/10.3390/ani13132020

\bibitem{Qiao2022}
Qiao, Y.; Guo, Y.; Yu, K.; He, D. C3D-ConvLSTM based cow behaviour classification using video data for precision livestock farming. \emph{Comput. Electron. Agric.} \textbf{2022}, \emph{193}, 106650. https://doi.org/10.1016/j.compag.2021.106650

\bibitem{Wu2023}
Wu, D.; Han, M.; Song, H.; Song, L.; Duan, Y. Monitoring the respiratory behavior of multiple cows based on computer vision and deep learning. \emph{J. Dairy Sci.} \textbf{2023}, \emph{106}, 2963--2979. https://doi.org/10.3168/jds.2022-22794

\bibitem{Maria2021}
Càmara, J.M.; Figueroa, S.; Rácz, F.; Duthoit, R.; Prieur, T.; Ranjith, R.; Crilly, E. Using imagery and computer vision as remote monitoring methods for early detection of respiratory disease in pigs. \emph{Comput. Electron. Agric.} \textbf{2021}, \emph{187}, 106283. https://doi.org/10.1016/j.compag.2021.106283

\bibitem{Wu2020}
Wu, D.; Wu, Q.; Yin, X.; Jiang, B.; Wang, H.; He, D.; Song, H. Lameness detection of dairy cows based on the YOLOv3 deep learning algorithm and a relative step size characteristic vector. \emph{Biosyst. Eng.} \textbf{2020}, \emph{189}, 150--163. https://doi.org/10.1016/j.biosystemseng.2019.11.013

\bibitem{Barney2023}
Barney, S.; Dlay, S.; Crowe, A.; Kyriazakis, I.; Leach, M. Deep learning pose estimation for multi-cattle lameness detection. \emph{Sci. Rep.} \textbf{2023}, \emph{13}, 4499. https://doi.org/10.1038/s41598-023-30962-7

\bibitem{Myint2024}
Myint, B.B.; Onizuka, T.; Tin, P.; Aikawa, M.; Kobayashi, I.; Zin, T.T. Development of a real-time cattle lameness detection system using a single side-view camera. \emph{Sci. Rep.} \textbf{2024}, \emph{14}, 13734. https://doi.org/10.1038/s41598-024-72436-8

\bibitem{Sohan2023}
Sohan, M.F.; Solaiman, M.; Hasan, M.A. A survey on deepfake video detection datasets. \emph{Indones. J. Electr. Eng. Comput. Sci.} \textbf{2023}, \emph{32}, 1168--1176. https://doi.org/10.11591/ijeecs.v32.i3.pp1168-1176

\bibitem{zhao2018}
Zhao, K.; Bewley, J.M.; He, D.; Jin, X. Automatic lameness detection in dairy cattle based on leg swing analysis with an image processing technique. \emph{Comput. Electron. Agric.} \textbf{2018}, \emph{148}, 226--236. https://doi.org/10.1016/j.compag.2018.03.005

\bibitem{jia2025}
Jia, Z.; Zhao, Y.; Mu, X.; Liu, D.; Wang, Z.; Yao, J.; Yang, X. Intelligent Deep Learning and Keypoint Tracking-Based Detection of Lameness in Dairy Cows. \emph{Vet. Sci.} \textbf{2025}, \emph{12}, 218. https://doi.org/10.3390/vetsci12020218

\bibitem{myint2024cattle}
Myint, B.B.; Zin, T.T.; Aikawa, M.; Kobayashi, I.; Tin, P. Cattle Lameness Detection Using Leg Region Keypoints from a Single RGB Camera. In \emph{International Conference on Genetic and Evolutionary Computing}; Springer Nature Singapore, 2024; pp. 180--189. https://doi.org/10.1007/978-981-97-2123-1\_17

\bibitem{tun2024}
Tun, S.C.; Onizuka, T.; Tin, P.; Aikawa, M.; Kobayashi, I.; Zin, T.T. Revolutionizing cow welfare monitoring: A novel top-view perspective with depth camera-based lameness classification. \emph{J. Imaging} \textbf{2024}, \emph{10}, 67. https://doi.org/10.3390/jimaging10030067

\end{thebibliography}

\end{document}